# Enhancing SHAP Explainability for Diagnostic and Prognostic ML Models in Alzheimer's Disease

**Pablo Guillén[1], Enrique Frias-Martinez[2,*]**

[1]MIU City University, Miami, FL, USA
[2]Research Institute UNIR iTED, Universidad Internacional de La Rioja (UNIR), Spain
[*]Corresponding Author: Enrique Frias-Martinez, enrique.frias@unir.net

**ABSTRACT:** Alzheimer's disease (AD) diagnosis and prognosis increasingly rely on machine learning (ML) models. Although these models provide good results, clinical adoption is limited by the need for technical expertise and the lack of trustworthy and consistent model explanations. SHAP (SHapley Additive exPlanations) is commonly used to interpret AD models, but existing studies tend to focus on explanations for isolated tasks, providing little evidence about their robustness across disease stages, model architectures, or prediction objectives. This paper proposes a multi-level explainability framework that measures the coherence, stability and consistency of explanations by integrating: (1) within-model coherence metrics between feature importance and SHAP, (2) SHAP stability across AD boundaries, and (3) SHAP cross-task consistency between diagnosis and prognosis. Using AutoML to optimize classifiers on the NACC dataset, we trained four diagnostic and four prognostic models covering the standard AD progression stages: normal-control (NC), mild-cognitive impairment (NCI) and AD. For each model, we generated SHAP and feature importance (FI) plots. Stability was then evaluated using correlation metrics (Spearman, Kendall), top-k feature overlap (Jaccard@10/20), SHAP sign consistency, and domain-level contribution ratios. Results show that cognitive and functional markers (e.g., MEMORY, JUDGMENT, ORIENT, PAYATTN) dominate SHAP explanations in both diagnosis and prognosis. SHAP-SHAP consistency between diagnostic and prognostic models was high across all classifiers ($\rho$ = 0.61–0.94), with 100% sign stability and minimal shifts in explanatory magnitude (mean Δ|SHAP| < 0.03). Domain-level contributions also remained stable, with only minimal increases in genetic features for prognosis. These results demonstrate that SHAP explanations can be quantitatively validated for robustness and transferability, providing clinicians with more reliable interpretations of ML predictions. The proposed framework provides a reproducible methodology for evaluating explainability stability and coherence, supporting the deployment of trustworthy ML systems in AD clinical settings.

## 1 Introduction

Alzheimer's disease (AD) is the most common form of dementia, impacting millions of people and their families [1]. This neurodegenerative disorder is characterized by cognitive decline and memory loss, and to date, there are no effective treatments capable of slowing down its progression. In this context, early detection of AD and its prior state, mild cognitive impairment (MCI), is a key element for defining the intervention and management of a patient.

Artificial intelligence techniques, and machine learning (ML) in particular, have emerged as relevant support tools, improving the precision and efficiency of AD diagnosis and prognosis. ML algorithms have been shown to be very efficient in distinguishing between normal controls (NC), Mild cognitive impairment (MCI), and Alzheimer's disease (AD) stages [2][3] as well as predicting progression between stages [4]. However, the adoption of these techniques in clinical settings is limited by their complexity, requiring specialized technical knowledge for training, deployment and interpretation. The black-box nature of many ML algorithms highlights the need for explainability to facilitate trust and integration into medical workflows.

Automated machine learning platforms (AutoML) have been introduced to reduce dependence and specialized ML experts. They offer user-friendly interfaces that enable non-technical personnel to develop models with a no-code or low-code approach. Explainability is key in medical contexts, and AutoML tools frequently integrate explainability algorithms to address this need, mainly SHapley Additive exPlanations (SHAP) and Local Interpretable Model-agnostic Explanations (LIME) [5][6].

SHAP has become one of the most common methods for AD. Nevertheless, the complexity of interpreting SHAP values, especially for non-technical clinicians, can itself become a barrier to understanding the results. Often these explanations focus only on individual models without assessing the stability, robustness and transferability of these interpretations across different models [7]. The conclusions obtained from this model-centric approach may not be applicable or consistent across different clinical scenarios. Finally, while SHAP is now widely used to interpret AD models, literature reviews indicate that there are heterogeneous reporting practices, scarce external validation, and limited assessment of explanation robustness [8].

This paper addresses the lack of systematic methods for evaluating the robustness and transferability of SHAP explanations in AD modeling. Existing studies on explanation robustness and stability typically evaluate robustness under data perturbation or sampling variability within a single task [9]. We propose a structured explainability framework that quantifies (i) coherence between feature importance and SHAP within models, (ii) stability of explanations across diagnostic scenarios, and (iii) transferability of explanations between diagnostic and prognostic tasks. By introducing multi-level stability metrics, the framework enables explainability to be treated as a measurable property rather than a purely qualitative element. This perspective extends explainability from model robustness to clinical robustness, enabling validation of whether the same explanatory structures are valid across disease stages and predictive objectives. This framework is designed to bridge the gap between ML models and the need for transparent, trustworthy, and stable explanations in clinical settings, allowing for a deeper understanding of the disease mechanisms and facilitating the identification of robust markers. Although our experiment focuses on AD, the idea of finding stable and trustworthy explainability is needed for any clinical application. As a result, the same approach can potentially be taken to improve the explainability of other illnesses.

To demonstrate the potential of AutoML for explainability in clinical applications, we employ PyCaret as a framework that provides SHAP as one of its Explainable AI (XAI) algorithms. PyCaret has proven effective in several medical applications such as thyroid disease, renal function, and diabetes. Two complementary tasks have been defined to study the stability and coherence of the explanation models produced by SHAP: (1) classification of cognitive impairment stages (NC, MCI, AD), i.e., diagnosis; and (2) prediction of future cognitive states four years after their initial visit, i.e., prognosis.

Our experiments use the National Alzheimer's Coordinating Center-Uniform Data Set (NACC-UDS) [10][11], a comprehensive and standardized resource employed in AD research. It focuses on clinical and behavioral variables. Previous studies have applied NACC to early detection, progression prediction, and diagnostic support for AD [12][13][14]. The paper is structured as follows. Section 2 presents related studies focusing on explainability for AD models. Section 3 describes the dataset, the methods used and the experimental design. Section 4 presents the results and evaluates the stability of the markers for AD and Section 5 discusses the results and compares them with state-of-the-art methods.

## 2 Related Studies

Explainable AI (XAI) focuses on methods to understand and trust the results of ML models [15][16]. There are two main approaches: (1) Intrinsic explainability, which implies the use of interpretable algorithms, like trees and their variants and (2) Post-hoc explainability, which refers to the application of explainability methods after model training. Post-hoc explainability methods can be further divided into: (1) model-specific techniques, which are restricted to specific algorithms; and (2) model-agnostic methods, which can be applied to any ML model. XAI plays a central role in AD because clinical tools require transparency and trust in decisions that affect diagnosis and treatment. XAI approaches for AD can be grouped into four main groups: (1) feature importance (FI), which is the most basic form of XAI, but still relevant for AD [17]; (2) SHAP (and its variants), which is the method most widely used because it measures both global and local feature contributions, enabling direct links between clinical markers and model predictions; (3) Grad-CAM, which focuses on spatial interpretability in neuroimaging by generating maps that highlight brain regions most influential in MRI or PET classifications; (4) LIME, which supports localized transparency for cognitive and demographic data by approximating complex models with linear solutions; and (5) Layer-wise relevance propagation (LRP), designed for CNNs, which traces how relevance flows through its layers. These methods can also be clustered depending on the type of data that they can be applied to, with SHAP and LIME mainly used for tabular data, while Grad-CAM and LRP are focused on MRI images.

SHAP groups a variety of algorithms based on game-theory that construct model-agnostic explanations of how each feature contributes to a prediction [18]. There are several SHAP implementations in the literature that adapt to different ML approaches: (1) KernelSHAP, a model-agnostic variant that calculates Shapley values through sampling and weighted linear regression, commonly applied to clinical or demographic data; (2) TreeSHAP designed for tree-based ensembles such as XGBoost, LightGBM, and Random Forests; (3) DeepSHAP which extends the method to deep neural networks allowing voxel- or feature-level interpretability and (4) GradientSHAP and PartitionSHAP that improve scalability and sensitivity to correlated features.

Focusing on AD, literature reviews show the adoption of SHAP for AD diagnosis and prognosis [7]. The main limitations reported are regarding the lack of standardization in how explanations are evaluated or presented to clinicians. When compared to other XAI methods, SHAP provides higher local fidelity than LIME [7]. Although in the context of AD SHAP is mainly applied to tabular data, some modifications can also apply to MRI and PET AD images [19].

The NACC-UDS dataset has been previously used to provide SHAP explainability for AD models. Some examples include reference [20], which implemented an XGBoost–SHAP model where Clinical Dementia Rating (CDR), Functional Activities Questionnaire (FAQ) and Mini-Mental State Examination (MMSE)showed the highest SHAP explainability levels. References [21][9] confirmed that MMSE, CDR, and FAQ items hold the largest global SHAP values for distinguishing AD stages. Table 1 summarizes examples found in the literature. Functional measures (FAQ) and cognitive assessments (CDR and MMSE) consistently receive high SHAP attribution in clinical models, highlighting their importance in model explainability. Genetic variant indicators show moderate importance, though general metadata flags (e.g., ExAC variant presence) show limited relevance due to sparse representation in the analyzed cohorts. Longitudinal changes show strong explanatory power in conversion models, indicating that SHAP can capture disease evolution.

Although existing SHAP-based studies have demonstrated the clinical relevance of cognitive and functional markers, they primarily rely on qualitative inspection of single-model explanations. Recent studies highlight that SHAP does not typically account for the stability and robustness of explanations across model architectures, sampling variations, or input perturbations which may produce inconsistent markers [22]. In this context there are studies that highlight the importance of introducing stability for explainability [23]. More broadly, reference [24] argues that XAI lacks metrics that meaningfully reflect explanation reliability, domain relevance, and task transferability.

Following this shift toward multi-dimensional and stable interpretability, this paper proposes a framework to evaluate explainability at three complementary levels: within-model coherence, cross-scenario stability, and cross-task transferability between diagnosis and prognosis. This enables quantitative validation of explanation robustness beyond the scope of previous approaches.

Table 1. SHAP-Based relevance found in the literature of NACC Feature Groups for AD.

| NACC Group | Relevance | References | Notes |
| --- | --- | --- | --- |
| **Functional Activities Questionnaire (FAQ)** | Yes | [2] [3] [22] [9] | FAQ items (BILLS, TAXES, PAYATTN, and TRAVEL) consistently appear among top SHAP-ranked features. Loss of function in instrumental daily activities is strongly predictive of AD classification. |
| **Clinical Dementia Rating (CDR) and MMSE** | Yes | [2] [3] [22] | CDR domains (e.g., MEMORY, ORIENT, JUDGMENT) and MMSE total score receive the highest SHAP values in diagnostic models. Lower MMSE and higher CDR scores are associated with increased AD probability. |
| **Genetic Flag** | Moderate | [9] | If available, APOE ε4 SHAP shows risk-modifying importance. |
| **GWAS Flag** | Moderate | [9] | If available SHAP highlights GWAS variants as secondary predictors. |
| **ExAC Flag** | Low/None | — | Rare-variant indicators seldom appear in published SHAP feature rankings due to sparse data. |
| **Progression** | Yes | [3] [9] | ΔMMSE, ΔCDR predicts MCI conversion |

## 3 Methods and Materials

### 3.1 Dataset

The data used in this study originated from the National Alzheimer's Coordinating Center Uniform Data Set (NACC-UDS). The NACC Uniform Data Set (NACC-UDS) provides longitudinal clinical information from Alzheimer's Disease Research Centers across the United States, including demographics, medical history, neuropsychological batteries, and clinician diagnoses [10][11]. It contains participant characteristics collected in Alzheimer's Disease Research Centers (ADRC) in the period 2005–2025. NACC-UDS includes demographic information, medical history, family history, medication usage, cognitive assessments, neurological exams, and clinical diagnoses. When accessed for this study, in March 2025, NACC contained 53,318 unique participants comprising 1,023 variables. Because NACC contains samples from baseline visits along with follow up visits of participants, the total number of instances are 195,196. Participants were classified using the NACC variable NACCUDSD, which indicates the level of cognitive impairment: NC, MCI or AD. The 1,024 features are divided into 7 main groups:

1. Functional Activities Questionnaire (FAQ). This set of features measures instrumental activities of daily living, including BILLS, which measures the ability to manage basic financial matters; TAXES, which measures the ability to handle complex financial matters; PAYATTN (Pay Attention), which evaluates attention and concentration; and TRAVEL that assesses the ability to travel independently.

2. Clinical Dementia Rating (CDR). CDR score measures the relative severity of dementia, ranging from 0 (no impairment) to 3 (severe impairment). It is determined by a clinician's judgment using an interview with the patient and a caregiver. Cognitive functions are assessed using a standardized neuropsychological battery, including the Mini-Mental State Examination (MMSE). CDR assesses dementia severity across several domains, including MEMORY, which evaluates short-term and long-term memory; ORIENT (Orientation) that measures awareness of time, place, and person;

JUDGMENT (Judgment and Problem Solving), that assesses decision-making, and COMMUN (Community Affairs) which refers to functioning in community settings.

(3) Indicators for Visit Packet: This feature (PACKET) identifies the type of visit packet used to collect data during a subject's evaluation. In the NACC dataset, different packet types correspond to the version of the Uniform Data Set (UDS) protocol implemented at the time of the visit. This feature is designed to distinguish between different versions of the protocol.

(4) Indicators for Language Codes: This feature (MOCALANX) specifies the language in which the Montreal Cognitive Assessment (MoCA) was administered. This is relevant for justifying variability in the results due to the patient's knowledge of the language in which the test was administered.

(5) Indicators for Genetic Data: This feature (ADGCEXR) identifies whether the patient's genetic data is available through the Alzheimer's Disease Genetics Consortium (ADGC). ADGC is a different initiative from NACC designed to collect genetic variants from AD. If available both profiles can be linked.

(6) Indicators for Presence of Genetic Variant: This feature (NGDSGWAC) indicates whether a patient´s genetic data includes variants identified through the Genome-Wide Association Study (GWAS). It basically indicates if the patient has known risk-associated genetic variants.

(7) Indicators for Presence of Genetic Variant (ExAC): This feature (NGDSEXAC) indicates the presence of genetic variants found in the Exome Aggregation Consortium (ExAC). It basically indicates if the patient has rare variants that may contribute to AD.

Finally, although NACC is one of the most comprehensive longitudinal AD datasets, it is not free of bias. Participants are recruited from AD Research Centers, which may over-represent some specific demographic profiles. This potential bias should be considered when interpreting both predictive performance and transferability of explainability results.

### 3.2 Data Pipeline

From the 1,024 features present in the NACC dataset, 882 correspond to numerical features and 142 correspond to categorical features. Starting from this dataset, a pipeline was designed to implement feature engineering and hyperparameter tuning. Table 2 presents the class distribution for both tasks and all scenarios after feature engineering. The following techniques were implemented:

(1) Missing values: Any feature with more than 50% of missing values was removed. This threshold is commonly used in ML, as features with excessive missing values tend to be unreliable and will degrade model performance [25]. This is especially relevant in medical applications [26]. For the rest of the features, missing values (if any) were assigned using the median value for continuous variables. Using the median value is less sensitive to outliers than using the mean [27]. For categorical values, mode imputation was used, i.e., the most frequently occurring category was assigned if missing. Using mode imputation ensured that the most representative category is retained.

(2) Scaling: For comparing features that are represented in different units, first we need to standardize them, typically centering around 0 with a standard deviation of 1. By scaling numerical variables, each feature contributes equally to the analysis, and the convergence of the ML algorithm is facilitated [28]. Standardization is especially important for clinical datasets, where measurement units vary significantly. As noted by [29] in the context of medical applications, scaling improves training stability and model accuracy. The following features of cognitive functioning were scaled to have a mean of 0 and a standard deviation of 1: MEMORY, ORIENT, JUDGMENT, COMMUN, BILLS, TAXES, PAYATTN, and TRAVEL.

(3) Encoding: This process converts categorial data into numerical format. For categorical features, excessive cardinality can lead to data sparsity and, as a result, encoding is recommended to ensure stable feature representation. High-cardinality features (defined as features with more than 50 unique categories) were encoded with a custom frequency encoder, in which each category was represented by the number of times it appeared in the data. Low-cardinality features (defined as features with less than 50 unique categories) were one-hot encoded, with each category having its own binary column. This approach avoids introducing non-existent ordinal relationships between categories and facilitates model interpretability [30]. Examples of encoded features included MOCALANX, ADGCEXR and NGDSGWAC. The naming of the feature indicates its encoding, with scaler_ indicating scaled features and ohe_ one-hot-encoded features.

No explicit feature selection or dimensionality reduction was applied beyond missing-value filtering and standard preprocessing to avoid introducing bias into the explainability analysis. All reported feature counts refer to the final feature space used by the models after preprocessing, including scaling of numerical variables and expansion of categorical variables via one-hot or frequency encoding.

**Table 2**. Class distribution for each task (diagnosis and prognosis). It includes the percentage of each label and the number of elements after preprocessing and before resampling.

| Task | Scenario | NC | MCI | AD |
| --- | --- | --- | --- | --- |
| **Diagnosis** | NC vs AD | 62.2% (94,933) | – | 37.8% (57,590) |
| | NC vs MCI | 73.5% (94,933) | 26.5% (34,106) | – |
| | MCI vs AD | – | 37.2% (34,106) | 62.8% (57,590) |
| | NC vs MCI vs AD | 50.8% (94,933) | 18.4% (34,196) | 30.8% (57,590) |
| **Prognosis** | NC vs AD | 45.6% (1,803) | – | 54.4% (2,150) |
| | NC vs MCI | 70.7% (1,803) | 29.3% (748) | – |
| | MCI vs AD | – | 25.8% (748) | 74.2% (2,150) |
| | NC vs MCI vs AD | 36.2% (1,803) | 15.0% (748) | 48.8% (2,150) |

## *3.3 AutoML*

Traditional ML environments require technical expertise, posing significant barriers to clinical adoption, especially in low-resource areas. With the increasing number of non-specialists working with ML applications, there has been an attempt to automate components of the ML workflow, introducing the concept of Automated Machine Learning (AutoML). AutoML tools automate model selection, hyperparameter tuning, and validation, allowing healthcare professionals to use ML techniques without extensive computational knowledge. There are a variety of AutoML platforms, such as AutoWEKA [31], auto-sklearn [32], Google AutoML [33], PyCaret [34] or AutoGluon [35]. In the literature, there is evidence that demonstrates the potential of AutoML platforms for improving ML model development when compared to manual approaches [36].

PyCaret [34] is an open-source, low-code machine learning library in Python that automates ML workflows. It offers over 25+ algorithms for supervised and unsupervised learning, and 70+ automated open-source algorithms. Examples of some of the algorithms included are: (1) GBC, or Gradient Boosting Classifier, which is the basic form of gradient boosting; (2) XGBoost, short for Extreme Gradient Boosting, an optimized version of GBC; (3) LightGBM, which takes XGBoost's approach a step further by speeding up the training process and reducing memory usage; (4) Extra Tree

Classifier, or ET, which is an ensemble learning technique that aggregates the results of multiple decision trees; and (5) Random Forest, RF, an ensemble learning method that constructs a multitude of decision trees. PyCaret is a good approach when using tabular data, such as in our case, and when some form of explainability is needed by the model. Tabular data is extremely common in medical applications, for example electronic health records or diagnostic results, which has caused PyCaret to be extensively used in this context [9].

Using PyCaret, two tasks have been defined, a diagnostic task and a prognostic task. The diagnostic task focuses on the classification of cognitive impairment stages, and the prognostic task targets the prediction of future cognitive states four years after their initial visit. Each task presents four scenarios, three binary (NC vs AD, CV vs. MCI, MCI vs AD) and one multiclass (NC vs. MCI vs AD). The prognostic task can be presented as a binary classification of progression vs non-progression; the output of the system is not a classification but an indication of the progression or non-progression of the illness. After the exploratory data analysis and feature selection, for the diagnostic task, NC vs AD used 37 features, as did NC vs. MCI and NC vs. MCI and AD, while MCI vs AD used 77 features. For the prognostic task, NC vs AD had 52 features NV vs MCI and MCI vs AD had 53 features and NC vs MCI vs AD had 54 features. Appendix A details the set of features selected for each task.

Feature size varies across scenarios due to three elements: (1) scenario-specific data availability of the NACC dataset; (2) filtering; and (3) task-dependent feature selection. Each classification scenario is constructed from a different subset of NACC, which leads to different distributions of missing values and categorical diversity. Additionally, the complexity of the clinical boundary influences the number of features retained by AutoML. As a result, well-separated stages (e.g., NC vs AD) are largely explained by a compact set of markers, producing smaller feature sets (≈37 features). In contrast, clinically adjacent stages (e.g., MCI vs AD) have greater overlap in symptomatology, requiring a larger number of feature sets (≈77 features). Prognostic scenarios show more consistent feature counts (≈52–54 features) because they rely on baseline variables that are longitudinally complete for subjects with follow-up visits. More tails can be found in [37].

For each scenario, an 80-20 train-test split was applied to hold out unseen data for final test evaluation. To prevent information leakage arising from longitudinal follow-up visits, all data splits were performed at the subject level rather than the visit level. Participant identifiers were used to group all visits from the same individual, ensuring that each subject appeared exclusively in either the training or the test set. This grouping constraint was maintained during cross-validation, so that no fold contained visits from a subject present in any other fold. After that, models were trained and hyperparameters optimized on the training set using 10-fold cross-validation with random grid search. SMOTE was applied within each training fold during cross-validation to ensure that synthetic samples did not affect the validation. Once the best model and parameter configuration were selected, the model was retrained on the full training data and evaluated on the 20% test set, providing an estimate of its generalization performance. Although SMOTE may introduce synthetic correlations, its application in this study was restricted to the training folds and was performed after subject-level splitting. No synthetic samples were included in validation or test sets.

### 3.3 Explainable AI (XAI)

Explainable AI (XAI) studies methods that allow users to understand and trust the results of a model. Pycaret includes, among others, two main tools for explainable AI, feature importance (FI) plots and SHAP. A feature importance plot is a post-hoc, model agnostic explainability method that quantifies how much each input variable contributes to the predictive performance of a ML model. The plot is typically built using permutation importance, an algorithm-agnostic approach that measures the change in model performance when the values of one feature are randomly changed while keeping all others constant. If permuting a feature reduces the model's performance, that feature is considered important. In the context of explainable medical AI, this plot offers interpretability by providing a ranking of variables by their overall influence on predictions.

SHAP [16] is a post-hoc model-agnostic explainability method that provides explanations to interpret the factors contributing to the model's decisions. SHAP assigns a weight, called the Shapley value, to each feature that represents its contribution to the prediction. The goal of SHAP is to explain the prediction of a sample $x_i$ by computing the influence score of each feature. The final prediction can be interpreted as the sum of the SHAP values. The prediction $y_i$ can be expressed as follows:

$$y_i = y_{base} + f(x_{i1}) + f(x_{i2}) + \cdots + f(x_{in}) \tag{1}$$

where $y_{base}$ is the average of predictions of all samples, and $f(x_{ij})$ is the SHAP value of $x_{ij}$ which is the contribution of feature $j$ to the prediction of $x_i$. When $f(x_{ij}) > 0$, feature $j$ will boost the prediction, otherwise, it has a negative effect. In SHAP plots, as implemented in PyCaret, each point in the plot represents one observation in the dataset; the horizontal axis shows the SHAP value, which quantifies how much that feature contributed to increasing or decreasing the model's output relative to the baseline prediction. Color provides extra information, with red points corresponding to high values, and blue to low values. The spatial distribution of colors along the SHAP axis indicates the directionality of a feature's effect. For example, if high-value (red) points cluster on the positive side of the axis, higher feature values increase the model's predicted probability of the positive class. If red points cluster on the negative side, higher values reduce that probability. This combined representation using color and position allows us to infer which features are most influential and how their values influence model decisions.

### 3.4 Experiment Design

Using AutoML, diagnosis and prognosis tasks will identify the best ML models for each scenario. In total, eight models will be identified, four models for diagnosis and four models for prognosis, for NC vs AD, NC vs MCI, MCI vs AD, and NC vs MCI vs AD. For each one of these models, an FI and a SHAP plot will be produced.

As has been highlighted in the literature, explainability is typically done for each model individually, which provides a limited view, as the conclusions obtained from single model may not be applicable or consistent across different scenarios. Analyzing explainability across different AD stages and between diagnostic and prognostic tasks will provide improved explainability, as cross-model evaluation will capture stability and consistency. Consistent explanations across diagnostic models would indicate that the algorithms identified robust and meaningful markers rather than artifacts of data partitioning or model architecture. Extending this evaluation to prognosis allows us to analyze whether the same markers that discriminate diagnosis also contribute to predicting progression.

While FI reflects the model's internal reliance on each variable, SHAP values quantify how those variables influence individual predictions. Integrating both representations bridges the gap between model structure and predictive behavior. The coherence between the explanations for each pair of models would indicate that the models rely on coherent predictors. Considering this, we introduce a multi-level framework that adds quantitative validation layers to standard SHAP-based interpretability. The framework consists of three components: (1) Within-Model coherence, which will evaluate the stability of the explainability of each model considering SHAP and FI; (2) inter-model coherence, which will evaluate the stability between different models for AD progression; and (3) cross-task coherence, which tests whether explanatory structures persist from diagnosis to prognosis. To evaluate the stability of model explainability the following metrics will be used: (1) Spearman rank correlation; (2) Jaccard overlap at top-k (Jaccard@10 and Jaccard@20); (3) SHAP sign consistency and

(4) Kendall (τ). If not otherwise stated, all explainability comparisons are computed on the intersection of features shared by the models being compared.

- Spearman's rank correlation (ρ) measures agreement between two rankings. This can be done between SHAP-SHAP or FI-SHAP. The FI–SHAP alignment evaluates the coherence between embedded (model-centric) and post hoc (explanation-centric) interpretability, i.e. weather features identified as most relevant during training are also the most relevant in post hoc explanations. This alignment is relevant for the top 10 features, as consistency among the most influential features would improve confidence. High ρ values indicate strong concordance, indicating stable and interpretable model behavior. Two Spearman approaches will be implemented: (1) standard Spearmen and (2) robust Spearman. Standard Spearman evaluates rank-order agreement across the full set of features and reflects interpretability similarity between the two methods. This approach penalizes cases where one method highlights predictors that the other does not. To complement this, we calculate a robust Spearman correlation using only the intersection of the top 10 features. This approach excludes non-overlapping features and focuses on the relative ordering of predictors included in the overlap.
- Jaccard overlap at top-k quantifies the proportion of shared top features between models. This can be applied between SHAP-SHAP or FI-SHAP.. A high Jaccard index indicates consensus among models. Jaccard@10 (J@10) and Jaccard@20 (J@20) will be implemented.
- SHAP sign consistency (%) measures the proportion of features maintaining the same direction indicating interpretative stability and consistency in SHAP-SHAP comparison. For each feature appearing in both explanations, we compare whether the SHAP values have the same sign. The sign consistency percentage is calculated as the proportion of shared features whose mean SHAP direction matches. High consistency is an indicator that not only are the same features important, but they also have their influence in the same direction, indicating stability. Lower values suggest that some features may change from positive to negative influence (or vice versa), indicating instability in the model's learned.
- Kendall τ. Kendall's τ provides a complementary perspective on rank agreement by evaluating pairwise concordance and discordance between two ordered lists. Whereas Spearman assesses how well the ordering of all features aligns, Kendall's τ examines every possible pair of features and measures the proportion for which both methods (FI and SHAP) agree on the ordering.
- Precision and recall. Precision measures the proportion of top 10 FI features that also appear among the top 10 SHAP features. Recall quantifies the proportion of top 10 SHAP features that are also present in the top 10 FI list. Together, these complementary metrics evaluate how well SHAP recovers the model's FI features and how well the model's internal ranking accounts for the features that exhibit the strongest post hoc influence.

For the third experiment, the diagnostic–prognostic comparison, additional metrics were used:

- Mean Δ|SHAP| captures changes in explanatory magnitude between prognosis and diagnostic tasks. For each feature present in both plots |SHAP| is calculated for diagnosis and prognosis as the mean absolute SHAP value of all samples. Then, Δ|SHAP| is obtained for each feature by comparing the diagnostic and prognostic values. Mean Δ|SHAP| is obtained by averaging the |SHAP| values of the common set of features. Δ|SHAP| provides a measure of stability, with larger values indicating greater shifts in explanatory magnitude across tasks.
- Feature-type contribution ratios (CDR, FAQ, Packet, Genetic) capture domain-level differences in explanatory weight by adding the absolute SHAP values for all features within each category and normalizing by the overall SHAP magnitude. Because Alzheimer's diagnosis and prognosis may rely on different combinations of markers, comparing contribution ratios reveals whether the model places different emphasis on the types of features when predicting current disease state versus future progression:

$$\text{Contribution}_{\text{domain}} = \frac{\sum |SHAP|_{\text{domain}}}{\sum |SHAP|_{\text{total}}} \quad (2)$$

The selected stability metrics have been chosen to capture complementary aspects of explanation robustness. Rank-based correlations (Spearman's $\rho$ and Kendall's $\tau$) quantify global agreement in feature ordering, while Jaccard@k evaluates overlap among the most influential predictors. Sign consistency captures directional agreement, ensuring that stable features exert influence in the same direction, and mean $\Delta|SHAP|$ measures magnitude shifts between tasks.

## 4 Results

Table 3 presents the best models identified by AutoML for the diagnostic task for each scenario (NC vs AD, NC vs MCI, MCI vs AD and NC vs MCI vs AD), including the algorithm that produced the best result, the accuracy, AUC, Kappa, and both the micro and macro precision, recall and F1. Class-specific metrics are given to indicate the performance of each group, where misclassifying a class (NC; MCI, AD) has distinct implications. Table 4 presents the best models identified by AutoML for the prognosis task and the same set of metrics. For the diagnostic task, the highest accuracy was obtained in the NC vs AD task using the XGBoost classifier, achieving an accuracy of 0.986, AUC of 0.996, and Cohen's Kappa of 0.973. For the prognosis task, the best result was also observed for NC vs AD using LightGBM, which achieved an accuracy of 0.926, AUC of 0.976, and Kappa of 0.873. These high-performance metrics ensure that interpretability analysis is based on clinically relevant models. In both tasks, best results are obtained for NC vs AD classification as, in AD evolution, those are the two extreme and best differentiated cases, while the consideration of MCI, reduces the metrics due to the unbalanced dataset. In general, when considering the results for both the diagnosis and prognosis experiments, it is important to consider the class prevalence of the NACC dataset. NC and AD categories are well represented but MCI labels are comparatively fewer. As a result, NC vs MCI and MCI vs AD are more challenging problems than NC vs AD. SMOTE was applied to mitigate this class imbalance, but oversampling does not completely replicate the complexity of the original data, especially for clinical datasets. In this context, both precision and recall should be interpreted considering the prevalence of the labels, where in general models achieve higher precision at the cost of recall in minority classes.

For each one of these models, FI and SHAP plots were generated by Pycaret for explainability purposes. Figure 1 presents, for the NC vs AD scenario, the FI and the SHAP plots for the diagnostic task (panels a and b) and for the prognostic task (panels c and d). Figures 2 through 4 present the same information for the rest of the scenarios NC vs MCI, MCI vs AD and NC vs MCI vs AD respectively.

**Table 3:** Diagnosis Task. Performance of the best diagnostic models identified for each classification task (NC vs AD, NC vs MCI, MCI vs AD, and NC vs MCI vs AD).

| Classifier | Algorithm | Accuracy | AUC | Kappa | Class | Precision | Recall | F1-score |
|---|---|---|---|---|---|---|---|---|
| **NC vs AD** | XGBoost | 0.986 | 0.998 | 0.972 | 0 (NC) | 0.987 | 0.989 | 0.988 |
| | | | | | 1 (AD) | 0.989 | 0.987 | 0.988 |
| | | | | | **Macro avg** | 0.988 | 0.988 | 0.988 |
| **NC vs MCI** | XGBoost | 0.913 | 0.956 | 0.826 | 0 (NC) | 0.920 | 0.916 | 0.918 |
| | | | | | 1 (MCI) | 0.927 | 0.920 | 0.918 |
| | | | | | **Macro avg** | 0.919 | 0.918 | 0.918 |
| **MCI vs AD** | XGBoost | 0.925 | 0.961 | 0.850 | 0 (MCI) | 0.926 | 0.943 | 0.934 |
| | | | | | 1 (AD) | 0.942 | 0.924 | 0.933 |
| | | | | | **Macro avg** | 0.934 | 0.934 | 0.933 |
| **NC vs MCI vs AD** | ET | 0.895 | 0.960 | 0.840 | 0 (NC) | 0.916 | 0.934 | 0.925 |
| | | | | | 1 (MCI) | 0.901 | 0.906 | 0.903 |
| | | | | | 2 (AD) | 0.987 | 0.962 | 0.974 |
| | | | | | **Macro avg** | 0.935 | 0.934 | 0.934 |

**Table 4:** Prognosis Task. Performance of the best prognostic models identified for each classification task (NC vs AD, NC vs MCI, MCI vs AD, and NC vs MCI vs AD).

| Classifier | Classifier | Accuracy | AUC | Kappa | Class | Precision | Recall | F1-score |
|---|---|---|---|---|---|---|---|---|
| **NC vs AD** | LightGBM | 0.926 | 0.976 | 0.873 | 0 (NC) | 0.938 | 0.971 | 0.954 |
| | | | | | 1 (AD) | 0.970 | 0.936 | 0.953 |
| | | | | | **Macro avg** | 0.945 | 0.908 | 0.946 |
| **NC vs MCI** | ET | 0.827 | 0.884 | 0.811 | 0 (NC) | 0.865 | 0.933 | 0.898 |
| | | | | | 1 (MCI) | 0.927 | 0.855 | 0.889 |
| | | | | | **Macro avg** | 0.871 | 0.827 | 0.891 |
| **MCI vs AD** | ET | 0.902 | 0.943 | 0.892 | 0 (MCI) | 0.966 | 0.994 | 0.974 |
| | | | | | 1 (AD) | 0.955 | 0.994 | 0.974 |
| | | | | | **Macro avg** | 0.919 | 0.992 | 0.974 |
| **NC vs MCI vs AD** | ET | 0.814 | 0.915 | 0.854 | 0 (NC) | 0.821 | 0.930 | 0.872 |
| | | | | | 1 (MCI) | 0.903 | 0.831 | 0.866 |
| | | | | | 2 (AD) | 0.996 | 0.942 | 0.968 |
| | | | | | **Macro avg** | 0.828 | 0.834 | 0.880 |

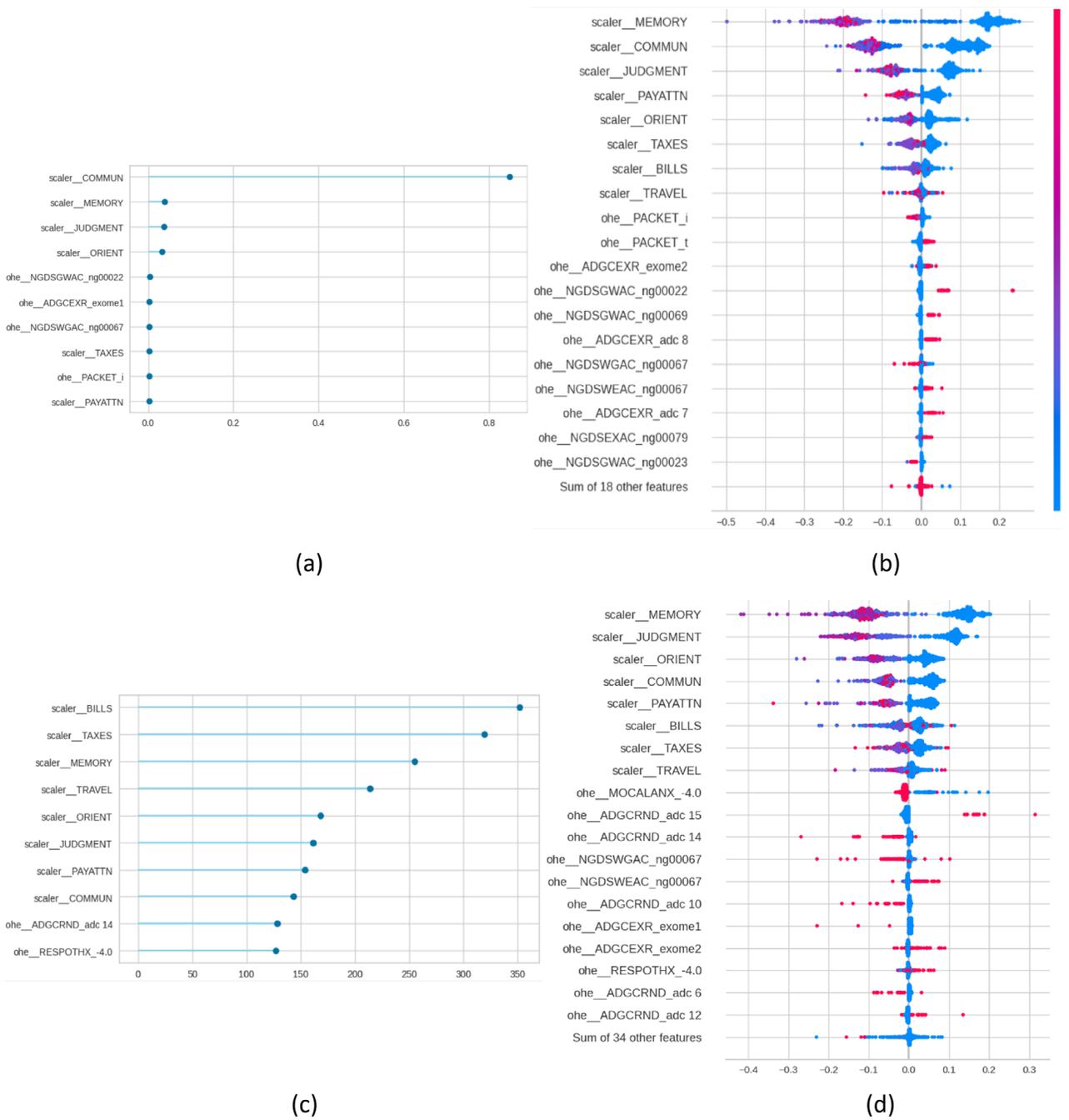

**Figure 1.** Feature Importance (FI) and SHAP plots for NC vs AD. Panels (a) and (b) present the plots for the diagnosis task, and panels (c) and (d) for the prognosis task.

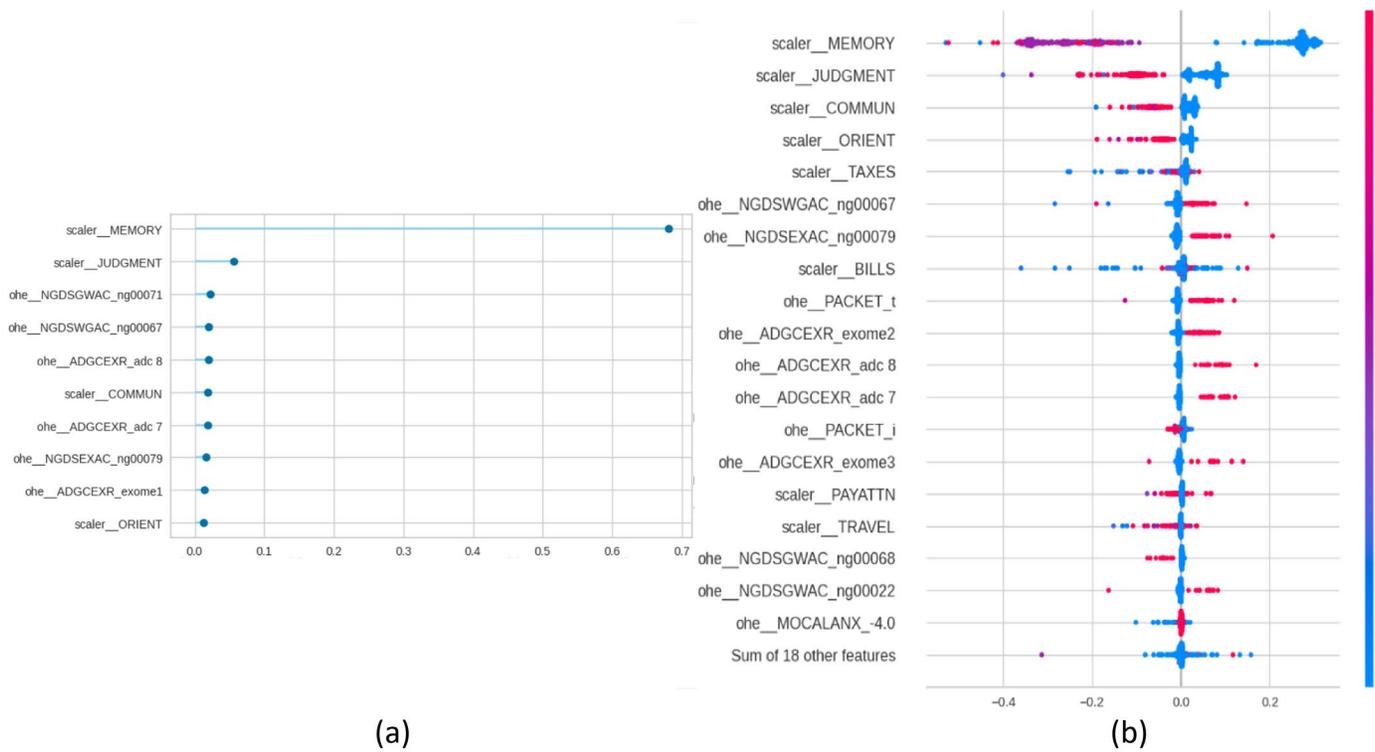
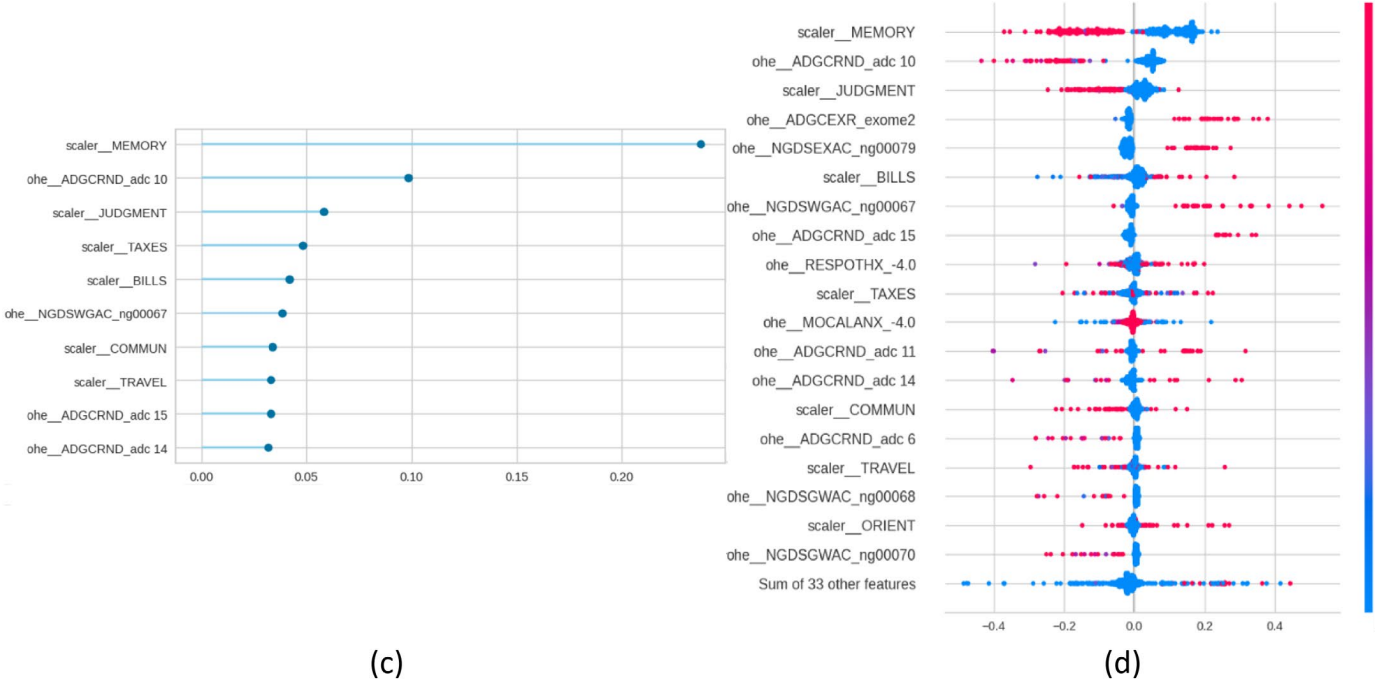

**Figure 2**. Feature Importance (FI) and SHAP plots for NC vs MCI. Panels (a) and (b) present the plots for the diagnosis task, and panels (c) and (d) for the prognosis task.

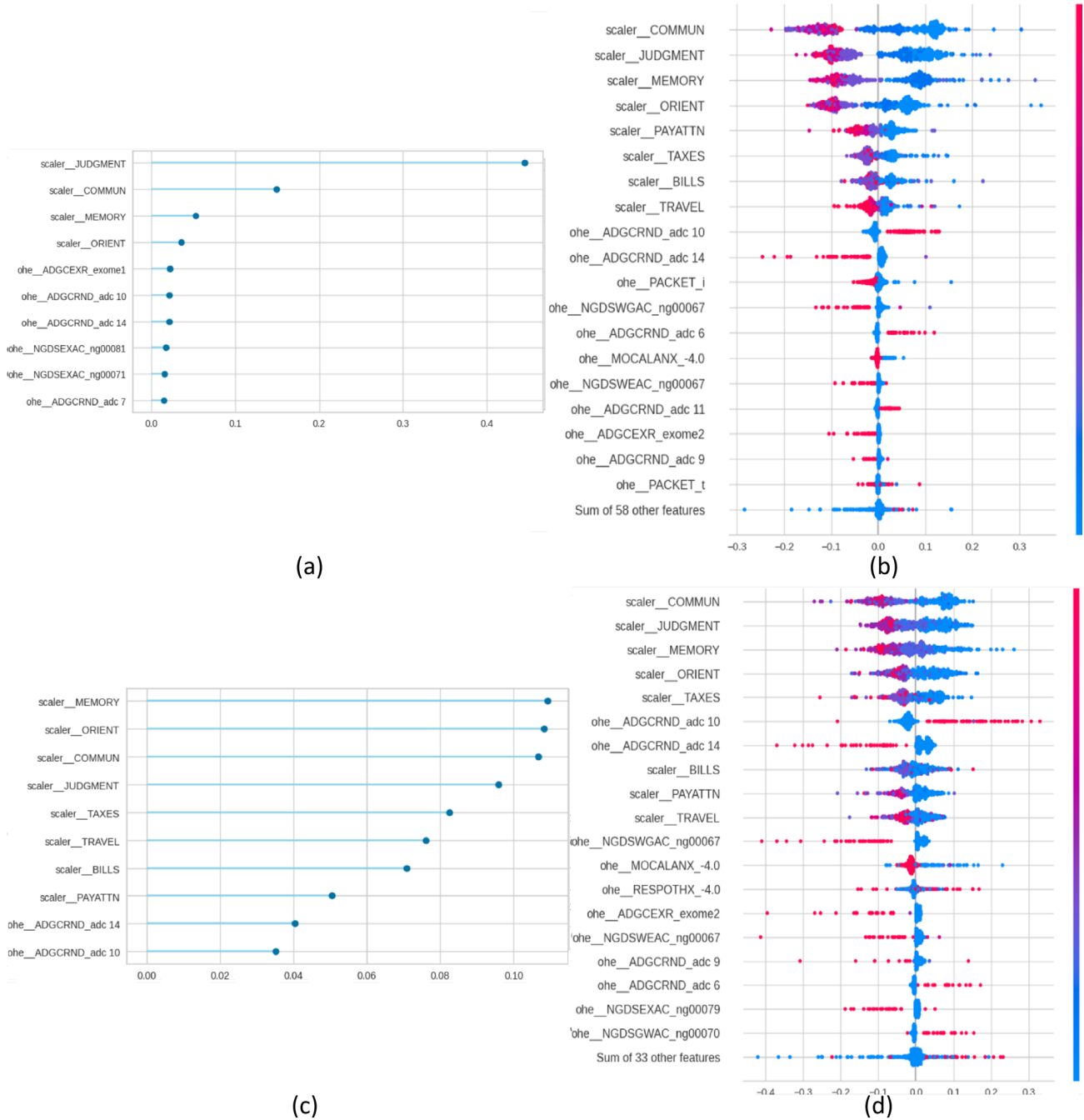

**Figure 3.** Feature Importance (FI) and SHAP plots for MCI vs AD. Panels (a) and (b) present the plots for the diagnosis task, and panels (c) and (d) for the prognosis task.

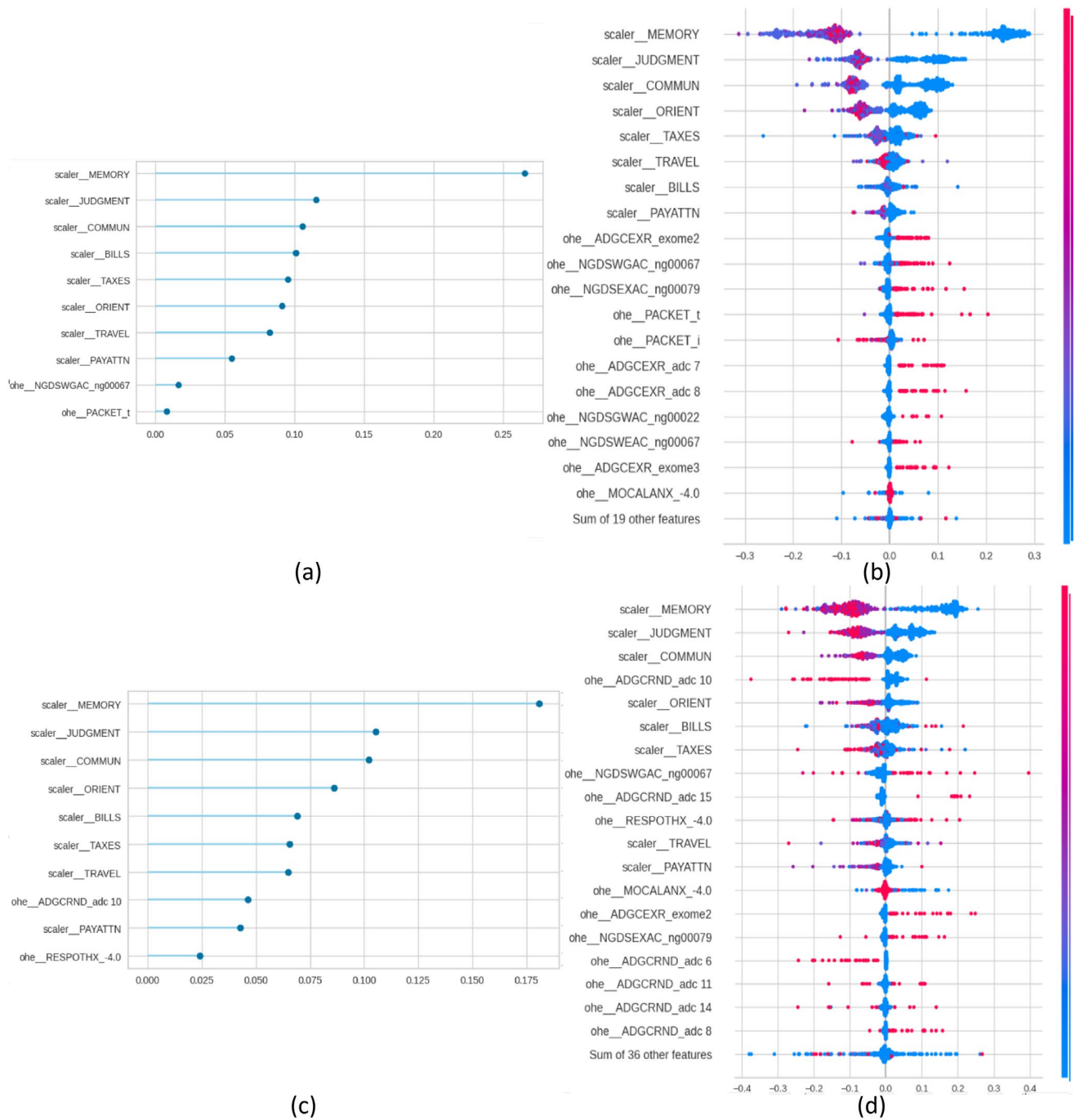

**Figure 4**. Feature Importance (FI) and SHAP plots for NC vs MCI vs AD. Panels (a) and (b) present the plots for the diagnosis task, and panels (c) and (d) for the prognosis task.

**Table 5.** Within-Model Explainability Coherence and Stability metrics. SHAP-FI metrics for each classifier and each task, including Spearman's rank correlation (ρ) and its robust variant (Robust ρ), Jaccard similarity at top 10 features (J@10), Kendall's Tau (τ), and Precision and Recall for top-ranked explanatory features.

| Classifier | Task | ρ | Robust ρ | J@10 | τ | Precission | Recall |
|---|---|---|---|---|---|---|---|
| **NC vs AD** | Diagnosis | 0.60 | 0.72 | 0.53 | 0.43 | 0.7 | 0.7 |
| | Prognosis | 0.75 | 0.75 | 0.70 | 0.56 | 0.8 | 0.8 |
| **NC vs MCI** | Diagnosis | 0.54 | 0.64 | 0.43 | 0.46 | 0.6 | 0.6 |
| | Prognosis | 0.67 | 0.94 | 0.43 | 0.83 | 1.0 | 1.0 |
| **MCI vs AD** | Diagnosis | 0.5 | 0.95 | 0.53 | 0.85 | 0.9 | 0.9 |
| | Prognosis | 0.67 | 0.87 | 1.00 | 0.72 | 0.8 | 0.8 |
| **NC vs MCI vs AD** | Diagnosis | 0.95 | 0.95 | 0.82 | 0.48 | 0.5 | 0.5 |
| | Prognosis | 0.76 | 0.95 | 0.67 | 0.83 | 1.0 | 1.0 |

**Table 6.** Within task classification Coherence and Stability Metrics. SHAP-SHAP metrics for diagnosis and prognosis classification, including Spearman's rank correlation (ρ), Jaccard similarity at top 10 features (J@10), Jaccard similarity at top 10 features (J@20), Kendall's Tau (τ), and sign consistency. Each cell indicates the metric value of the corresponding pairwise comparison to evaluate the consistency and stability of explanations factors when advancing the AD stage.

| | Diagnosis | | | | Prognosis | | | |
|---|---|---|---|---|---|---|---|---|
| Pairwise Comparison | ρ | J@10 | J@19 | τ | ρ | J@10 | J@20 | τ |
| **NC vs AD ↔ NC vs MCI** | 0.49 | 0.53 | 0.72 | 0.38 | -0.14 | 0.25 | 0.60 | -0.12 |
| **NC vs AD ↔ MCI vs AD** | 0.92 | 0.66 | 0.52 | 0.79 | 0.79 | 0.66 | 0.68 | 0.61 |
| **NC vs AD ↔ NC vs MCI vs AD** | 0.84 | 0.66 | 0.81 | 0.66 | 0.66 | 0.53 | 0.72 | 0.53 |
| **NC vs MCI ↔ MCI vs AD** | 0.69 | 0.42 | 0.52 | 0.49 | 0.28 | 0.33 | 0.72 | 0.13 |
| **MCI vs AD ↔ NC vs MCI vs AD** | 0.85 | 0.66 | 0.58 | 0.64 | 0.8 | 0.53 | 0.72 | 0.63 |

From a traditional perspective of explainability, in which each SHAP and FI graphs are treated individually, the figures indicate that cognitive tests are top predictors in all scenarios: MEMORY, JUDGMENT, COMMUN, ORIENT, and PAYATTN. The contribution of genetic markers (ADGCEXR_exome2, ADGCEXR_adc 7) and clinical evaluations (MOCALEXANX__4.0) was lower compared to cognitive assessments. In the multiclass classification task models showed a broader distribution of influential features compared to binary classifications. These findings are consistent with the state of the art, which has identified the core role of cognitive testing in Alzheimer's classification [3], and the value of genetic information for providing additional value for prognosis [3]. In any case, while SHAP indicates the statistical contribution of features to predictions, their clinical interpretability depends on domain expertise.

These conclusions, obtained from single models, may not be applicable or consistent across different clinical scenarios. The multi-level framework analysis of these SHAP and FI graphs is presented in Table 5, Table 6 and Table 7. Table 5 presents within-model coherence and stability metrics. This set of metrics reflects the agreement between algorithm-embedded feature importance (FI) and SHAP-derived importance for each of the eight models identified. FI and SHAP values represent two complementary layers of interpretability: FI quantifies the model's logic by measuring how much each feature contributes to reducing prediction error, whereas SHAP captures the influence of those variables on the model's outputs. Comparing these two forms allows us to evaluate if the model's structural reliance aligns with its observed predictive behavior. A high correlation between FI and SHAP indicates that the model's decision-making process is coherent, whereas a low correlation may reveal instability or redundant predictors. Results indicate a medium to high degree of coherence between FI and SHAP, with Spearman's ρ ranging from 0.50 to 0.95 and its robust variant showing greater agreement (0.64–0.95). This suggests a stable monotonic relationship between ranked importance values of FI and SHAP. The Jaccard similarity at the top 10 explanatory features (J@10) varies from 0.53 to 1.00, indicating that the overlap in dominant predictors depends on the classification task and disease stage. Kendall's τ values (0.43–0.85) further confirm moderate concordance in rank order. Precision and recall values between 0.6 and 1.0 show that, for most models, most FI-identified predictors were also recovered by SHAP. Note that precision and recall have the same values because they have been calculated using the same number of features. Within-model explainability results establish that SHAP explanations align closely with internal model importance rankings (FI), particularly in prognostic models. This suggests that SHAP provides coherent explanations that reflect the model's learned priorities. However, these results alone do not clarify whether explanations are resilient to variations in clinical stage or predictive objective. This is presented in Table 6 and Table 7.

**Table 7.** Cross-Task Explainability Coherence and Consistency (Diagnosis vs Prognosis) metrics. Each cell of the table indicates first the value for the FI-FI metric and the SHAP-SHAP metric for each classifier comparing the diagnosis and prognosis results. Note that J@20, Sign Consistency, Contribution and Mean Δ|SHAP| only apply to SHAP-SHAP comparison.

| Scenario | ρ | τ | J@10 | J@20 | Sign (%) | Contribution Diagnosis | Contribution Prognosis | Mean Δ|SHAP| |
|---|---|---|---|---|---|---|---|---|
| NC vs AD | -0.26 / 0.91 | -0.20/0.78 | 0.4/0.66 | -/0.40 | 1.0 | (0.47, 0.27, 0.08, 0.16) | (0.59, 0.25, 0, 0.15) | 0.022 |
| NC vs MCI | 1.0/0.39 | 1.0/0.24 | 0.4/0.53 | -/0.46 | 1.0 | (0.68, 0.21, 0, 0.14) | (0.62, 0.18, 0, 0.19) | 0.019 |
| MCI vs AD | -0.6 /0.87 | -0.67/0.86 | 0.4/1.0 | -/0.72 | 1.0 | (0.51, 0.17, 0.06, 0.25) | (0.38, 0.18, 0, 0.33) | 0.012 |
| NC vs MCI vs AD | 0.93 /0.61 | 0.93/0.4 | 0.8/0.70 | -/0.46 | 1.0 | (0.44, 0.27, 0.07, 0.22) | (0.38, 0.27, 0, 0.35) | 0.018 |

    Table 6 reports SHAP–SHAP agreement across disease stage scenarios and, as a result, evaluates the internal consistency across disease stages in diagnostic and prognostic tasks. It complements the within-model analysis by showing intra-task coherence, i.e., how stable feature explanations are across different scenarios. The objective is to test whether models trained on different clinical boundaries rely on the same markers. High similarity implies the models capture a coherent pattern; low similarity would indicate stage-specific features. Results show that diagnosis models exhibit high SHAP stability in most comparisons, with Spearman's ρ values exceeding 0.84 and J@10 above 0.60 in three of five comparisons. The NC vs AD ↔ MCI vs AD comparison shows the strongest agreement (ρ = 0.92, τ = 0.79), suggesting overlapping features across early and late AD stages. In contrast, NC vs AD ↔ NC vs MCI has the lowest agreement (ρ = 0.49), indicating differences between MCI and AD markers. Prognostic models exhibit more variability, particularly in early-stage transitions. NC vs AD ↔ NC vs MCI yielded a negative ρ (–0.14) and low Jaccard overlap (J@10 = 0.25), suggesting that progression models in early-stage rely on distinct markers. Negative correlations observed in early-stage prognostic transitions should be interpreted with caution, as they reflect the heterogeneity and

weak signal characterizing early disease progression rather than instability of the explainability framework. Consistency improves in comparisons involving MCI vs AD and multiclass settings, with ρ values ≥ 0.66 and J@10 ≥ 0.53. Diagnosis-stage pairs show J@10 scores up to 0.53 and sign consistency exceeding 60%, while prognosis models perform better, reaching J@10 = 0.82 and 100% sign agreement in multiple comparisons. These findings suggest that as AD progresses, the dominant predictors evolve. Prognosis comparisons show lower values in early transitions (e.g., NC vs AD ↔ NC vs MCI), indicating that robustness is context-sensitive: more stable in models that capture established disease markers and more variable in early-stage predictive tasks.

Table 7 presents cross-task coherence comparing diagnosis vs prognosis models. Each cell of the table indicates first the FI-FI metric and then the SHAP-SHAP metric between the diagnosis and prognosis models being compared. Focusing on SHAP-SHAP, two more columns are calculated: Contribution and Δ|SHAP|. Contribution provides two vectors, one for the diagnosis task and another for the prognosis task, where each number corresponds to the weight of CDR, FAQ, Packet, and Genetic features, respectively. By comparison of the two vectors, we can evaluate whether there is a shift in the importance of the group of features from diagnosis to prognosis. Results show that SHAP-SHAP rank correlation and agreement are moderate to high across classifiers, with Spearman ρ ranging from 0.61 to 0.94 and J@10 between 0.53 and 1.0. Kendall's τ values confirm moderate to strong consistent feature ordering. Sign consistency is uniformly 1.0, indicating that all shared features have the same directional contribution. The contribution vectors reveal small changes: CDR and FAQ maintain influence across tasks, but Packet and Genetic features slightly increase their weight in prognostic models. These shifts suggest that genetic and administrative context gains predictive value in longer-term progression modeling. The mean Δ|SHAP| values for top features are consistently low (0.012–0.022), reinforcing that the magnitude of explanatory change between diagnosis and prognosis remains minimal. These findings indicate that the same core predictors used for diagnosis also inform prognosis.

## 5 Discussion

From a clinical perspective, consistency of explanations across diagnostic and prognostic is critical for trust. Stable explanatory patterns across tasks indicate that the models rely on coherent clinical markers, enabling clinicians to interpret predictions consistently across scenarios. Prior studies, summarized in Table 1, identified cognitive and functional variables as primary contributors to SHAP-based explanations for AD models [2][3][17]. Additionally, genetic markers were identified as having moderate importance [9]. Although most of the predictors correspond to cognitive and functional markers, some features such as PACKET indicators reflect administrative artifacts rather than disease mechanisms. Their stability across tasks highlights consistency in data collection practices, not necessarily clinical relevance. In any case, our results confirm the state of the art, while at the same time extend the insight by demonstrating that these predictors are not only top-ranked in individual models, but, to a large extent, also remain stable across a variety of contexts: within models (Table 5), across clinical stages (Table 6), and between diagnostic and prognostic objectives (Table 7).

The literature shows that most of the existing research relies on qualitative inspection of individual SHAP plots. The framework presented in this paper shows that explanation robustness is a quality that can be quantified. From a methodological perspective, the integration of coherence metrics, such as ρ, τ, J@k, sign consistency, and Δ|SHAP|, with SHAP and FI plots enables explainability to be treated as a measurable property. This framework provides four advantages: (1) stability, as demonstrated by consistent rankings and directions across tasks and models; (2) coherence, as SHAP reflects embedded model logic; (3) transferability, shown by persistence across diagnostic and prognostic tasks; and (4) summarization, allowing comparative and scalable evaluation across scenarios and tasks. This framework can be applied to other clinical domains where explainability should be evaluated across multiple dimensions before being considered clinically trustworthy.

The combined view of Tables 5 through 7 presents a set of views that complement each other to define explainability. Table 5 shows that while prognostic models show stronger SHAP-FI alignment than diagnostic ones, this does not guarantee consistency across disease boundaries. Table 6 complement the previous result by evaluating within-task transitions, showing that SHAP explanations are more stable in mid- and late-stage comparisons than in early-stage ones. This suggests that while SHAP features remain constant within models, they shift during AD progression. Table 7 shows that the features identified for AD diagnosis transfer to prognostic models. When comparing diagnosis and prognosis tasks, SHAP rankings remain consistent (ρ = 0.87–0.94), directional contributions are preserved (sign = 100%), and domain weights shift only marginally. Also, the minimal Δ|SHAP| observed between diagnostic and prognostic models indicates that future cognitive status is mainly predicted from baseline impairment severity, reflecting the autoregressive nature of AD. While this limits the ability of the framework to capture AD progression, it does not diminish its relevance for near-term risk classification, which is grounded in baseline cognitive assessment. The prognostic task in this study is designed as an interpretability transfer experiment, assessing whether explanatory markers for diagnosis remain stable when predicting future cognitive status. Although predictions are driven by cognitive severity, this behavior reflects the autoregressive nature of AD rather than a limitation of the explainability framework

Finally, while our results demonstrate strong stability of SHAP explanations across diagnostic and prognostic tasks, these findings should not be interpreted as universal task-independent properties of SHAP. The current framework is limited to structured clinical variables and does not incorporate multimodal information such as neuroimaging.

## 6 Conclusions

The integration of ML models for AD diagnosis and prognosis depends on the ability to provide trustworthy, stable, and quantitative explanations. Previous SHAP-based AD studies, although building models that provided excellent diagnosis and prognosis metrics, rarely evaluated the stability of their explanations across models or tasks. Most existing work focuses on isolated diagnostic scenarios and provides explanations via visual inspection of SHAP plots. While those studies confirm the clinical relevance of SHAP-ranked features, they lack quantitative assessments of coherence.

This paper extends the state of the art by providing a structured evaluation of stability in explainability. By combining an AutoML pipeline with a multi-level explainability analysis, the paper has presented a framework to improve SHAP based explainability with cross-task metrics of stability and coherence. The proposed framework enables clinicians to assess whether model explanations are robust across disease stages and predictive objectives by providing quantitative evidence of explanation coherence and transferability, supporting more reliable interpretation of ML outputs.

The experiments were built using the NACC dataset to optimize AutoML models and quantify the coherence and robustness of their explanations. Diagnostic models achieved accuracy of 0.98 and AUC of 0.99 for the NC vs AD, and prognostic models achieved slightly lower metrics, with the NC vs AD achieving an accuracy of 0.92 and AUC of 0.97. Macro F1-scores exceeded 0.88 in all models, indicating balanced classification performance across AD stages. The enhanced explainability analysis was implemented in three levels: within model stability, intra scenario stability within the same task, and inter-task stability. Results indicate that SHAP-based explanations are both stable and coherent. Within-model explainability agreement between SHAP values and FI rankings was moderate to strong (Spearman ρ range: 0.50–0.95), with prognostic models exhibiting consistently higher SHAP-FI alignment than diagnostic models. Cross-task explainability (diagnosis vs prognosis) confirmed high interpretive consistency. The average change in SHAP magnitude was minimal (Δ|SHAP| < 0.03),

indicating that key predictors were not only preserved in rank but also retained their relative explanatory strength.

For future research, it will be relevant to extend this framework to study the stability of multimodal data sources, such as the combination of NACC dataset with MRI or PET images. Incorporating multimodal data represents the next step to evaluate if the proposed explainability framework generalizes to hybrid clinical–imaging models. Also, implementing longitudinal SHAP analysis will be key to assessing how feature importance and stability evolve over the course of the disease to provide more dynamic and temporally relevant explanations. Finally, an expansion of the multi-level stability metrics to include other XAI methodologies, such as LIME, would provide a comprehensive, comparative assessment of interpretability robustness.

**Acknowledgement:** The NACC database is funded by NIA/NIH Grant U24 AG072122. NACC data are contributed by the NIA-funded ADRCs: P30 AG062429 (PI James Brewer, MD, PhD), P30 AG066468 (PI Oscar Lopez, MD), P30 AG062421 (PI Bradley Hyman, MD, PhD), P30 AG066509 (PI Thomas Grabowski, MD), P30 AG066514 (PI Mary Sano, PhD), P30 AG066530 (PI Helena Chui, MD), P30 AG066507 (PI Marilyn Albert, PhD), P30 AG066444 (PI David Holtzman, MD), P30 AG066518 (PI Lisa Silbert, MD, MCR), P30 AG066512 (PI Thomas Wisniewski, MD), P30 AG066462 (PI Scott Small, MD), P30 AG072979 (PI David Wolk, MD), P30 AG072972 (PI Charles DeCarli, MD), P30 AG072976 (PI Andrew Saykin, PsyD), P30 AG072975 (PI Julie A. Schneider, MD, MS), P30 AG072978 (PI Ann McKee, MD), P30 AG072977 (PI Robert Vassar, PhD), P30 AG066519 (PI Frank LaFerla, PhD), P30 AG062677 (PI Ronald Petersen, MD, PhD), P30 AG079280 (PI Jessica Langbaum, PhD), P30 AG062422 (PI Gil Rabinovici, MD), P30 AG066511 (PI Allan Levey, MD, PhD), P30 AG072946 (PI Linda Van Eldik, PhD), P30 AG062715 (PI Sanjay Asthana, MD, FRCP), P30 AG072973 (PI Russell Swerdlow, MD), P30 AG066506 (PI Glenn Smith, PhD, ABPP), P30 AG066508 (PI Stephen Strittmatter, MD, PhD), P30 AG066515 (PI Victor Henderson, MD, MS), P30 AG072947 (PI Suzanne Craft, PhD), P30 AG072931 (PI Henry Paulson, MD, PhD), P30 AG066546 (PI Sudha Seshadri, MD), P30 AG086401 (PI Erik Roberson, MD, PhD), P30 AG086404 (PI Gary Rosenberg, MD), P20 AG068082 (PI Angela Jefferson, PhD), P30 AG072958 (PI Heather Whitson, MD), P30 AG072959 (PI James Leverenz, MD).

**Funding Statement:** E. Frias-Martinez would like to thank the IBM-UNIR Chair on Data Science in Education and the Research Institute for Innovation and Technology in Education (UNIR iTED) for partially funding this research.

**Author Contributions:** The authors confirm contribution to the paper as follows: Conceptualization, E. Frias-Martinez; methodology, Enrique Frias-Martinez; software, Pablo Guillén; formal analysis, Enrique Frias-Martinez, Enrique Frias-Martinez; investigation, Enrique Frias-Martinez, Pablo Guillén; data curation, Pablo Guillén; writing - original draft preparation Enrique Frias-Martinez; writing - review and editing, Enrique Frias-Martinez , Pablo. Guillén; visualization Pablo Guillén. All authors reviewed the results and approved the final version of the manuscript.

**Availability of Data and Materials:** The data supporting the findings of this study are available upon request from the National Alzheimer's Coordinating Center (NACC). The dataset used in this study was accessed in March 2025. For more information or to submit a data access request, please visit: https://naccdata.org/requesting-data/submit-data-request.

**Ethics Approval:** This study did not involve the recruitment of new human subjects. All analyses were conducted using secondary, de-identified data obtained from the National Alzheimer's Coordinating Center Uniform Data Set (NACC-UDS). Data collection and participant consent procedures were originally approved by the Institutional Review Boards (IRBs) of the contributing Alzheimer's Disease Research Centers. Access to the NACC-UDS was granted following approval of a formal data request by NACC. No additional ethical approval was required for this secondary analysis of anonymized data.

**Conflicts of Interest:** The author(s) declare no conflicts of interest to report regarding the present study.

## Appendix A

**Table 8.** Features used to build diagnostic task scenarios NC vs AD and NC vs MCI, grouped by type. Both scenarios use n=37 features.

| Cognitive / Functional | Packet/ Language | ADGCEXR | NGDSGWAC | NGDSEXAC | NGDSW-GAC/WEAC |
|---|---|---|---|---|---|
| MEMORY | PACKET_i | adc 7 | ng00022 | ng00071 | ng00067 |
| ORIENT | PACKET_it | adc 8 | ng00023 | ng00079 | ng00067 |
| JUDGMENT | PACKET_t | exome1 | ng00024 | ng00081 | |
| COMMUN | MOCALANX_-4.0 | exome1, adc 8 | ng00068 | | |
| BILLS | MOCALANX_2 | exome2 | ng00069 | | |
| TAXES | MOCALANX_cantonese | exome3 | ng00070 | | |
| PAYATTN | MOCALANX_english | | ng00071 | | |
| TRAVEL | MOCALANX_french | | | | |
| | MOCALANX_mandairin | | | | |
| | MOCALANX_mandarin | | | | |
| | MOCALANX_spanish | | | | |

**Table 9.** Features used to build diagnostic task scenario NC vs MCI vs AD, with n=38 features.

| Cognitive/ Functional | Packet/ Language | ADGCEXR | NGDSGWAC | NGDSEXAC | NGDSW-GAC/WEAC |
|---|---|---|---|---|---|
| MEMORY | PACKET_i | adc 7 | ng00022 | ng00071 | ng00067 |
| ORIENT | PACKET_it | adc 8 | ng00023 | ng00079 | ng00067 |
| JUDGMENT | PACKET_t | exome1 | ng00024 | ng00081 | |
| COMMUN | MOCALANX_-4.0 | exome1, adc 8 | ng00068 | | |
| BILLS | MOCALANX_0 | exome2 | ng00069 | | |
| TAXES | MOCALANX_2 | exome3 | ng00070 | | |
| PAYATTN | MOCALANX_cantonese | | ng00071 | | |
| TRAVEL | MOCALANX_english | | | | |
| | MOCALANX_french | | | | |
| | MOCALANX_mandairin | | | | |
| | MOCALANX_mandarin | | | | |
| | MOCALANX_spanish | | | | |

**Table 10**. Features used to build prognostic task scenario NC vs AD grouped by type, with n=52 features.

| Cognitive / Functional | Language | ADGCRND | ADGCEXR | NGDSGWAC | NGDSEXAC | NGDSWGAC/WEAC |
|---|---|---|---|---|---|---|
| MEMORY | PACKET_it | adc 1 | adc 7 | ng00022 | ng00071 | NGDSWGAC_ng00067 |
| ORIENT | MOCALANX_-4.0 | adc 10 | adc 8 | ng00023 | ng00079 | NGDSWEAC_ng00067 |
| JUDGMENT | RESPOTHX_-4.0 | adc 10, adc 11 | exome1 | ng00024 | ng00081 | |
| COMMUN | | adc 10, adc 12 | exome2 | ng00068 | | |
| BILLS | | adc 10, adc 14 | exome3 | ng00069 | | |
| TAXES | | adc 11 | | ng00070 | | |
| PAYATTN | | adc 11, adc 10 | | ng00071 | | |
| TRAVEL | | adc 11, adc 9 | | | | |
| | | adc 12 | | | | |
| | | adc 13 | | | | |
| | | adc 13, adc 15 | | | | |
| | | adc 14 | | | | |
| | | adc 14, adc 10 | | | | |
| | | adc 15 | | | | |
| | | adc 2 | | | | |
| | | adc 3 | | | | |
| | | adc 4 | | | | |
| | | adc 5 | | | | |
| | | adc 6 | | | | |
| | | adc 7 | | | | |
| | | adc 8 | | | | |
| | | adc 9 | | | | |
| | | adc 9, adc 5 | | | | |
| | | aa june2011 | | | | |

**Table 11.** Features used to build diagnostic task scenario NC vs MCI and scenario MCI vs AD, with n=53 features.

| Cognitive / Functional | Language | ADGCRND | ADGCEXR | GDSGWAC | NGDSEXAC | NGDSWGAC/WEAC |
|---|---|---|---|---|---|---|
| MEMORY | PACKET_it | adc 1 | adc 7 | ng00022 | ng00071 | NGDSWGAC_ng00067 |
| ORIENT | MOCALANX_-4.0 | adc 10 | adc 8 | ng00023 | ng00079 | NGDSWEAC_ng00067 |
| JUDGMENT | MOCALANX_cantonese | adc 10, adc 11 | exome1 | ng00024 | ng00081 | |
| COMMUN | RESPOTHX_-4.0 | adc 10, adc 14 | exome2 | ng00068 | | |
| BILLS | | adc 11 | exome3 | ng00069 | | |
| TAXES | | adc 11, adc 10 | | ng00070 | | |
| PAYATTN | | adc 11, adc 9 | | ng00071 | | |
| TRAVEL | | adc 12 | | | | |
| | | adc 12, adc 10 | | | | |
| | | adc 13 | | | | |
| | | adc 13, adc 15 | | | | |
| | | adc 14 | | | | |
| | | adc 15 | | | | |
| | | adc 2 | | | | |
| | | adc 3 | | | | |
| | | adc 4 | | | | |
| | | adc 5 | | | | |
| | | adc 6 | | | | |
| | | adc 7 | | | | |
| | | adc 8 | | | | |
| | | adc 9 | | | | |
| | | aa june2011 | | | | |

**Table 12.** Features used to build diagnostic task scenario NC vs MCI vs AD, grouped by type, with n=54 features.

| Cognitive / Functional | Language | ADGCRND | ADGCEXR | GDSGWAC | NGDSEXAC | NGDSWGAC/WEAC |
|---|---|---|---|---|---|---|
| MEMORY | PACKET_it | adc 1 | adc 7 | ng00022 | ng00071 | NGDSWGAC_ng00067 |
| ORIENT | MOCALANX_-4.0 | adc 10 | adc 8 | ng00023 | ng00079 | NGDSWEAC_ng00067 |
| JUDGMENT | MOCALANX_cantonese | adc 10, adc 11 | exome1 | ng00024 | ng00081 | |
| COMMUN | RESPOTHX_-4.0 | adc 10, adc 12 | exome2 | ng00068 | | |
| BILLS | | adc 10, adc 14 | exome3 | ng00069 | | |
| TAXES | | adc 11 | | ng00070 | | |
| PAYATTN | | adc 11, adc 10 | | ng00071 | | |
| TRAVEL | | adc 11, adc 9 | | | | |
| | | adc 12 | | | | |
| | | adc 12, adc 10 | | | | |
| | | adc 13 | | | | |
| | | adc 13, adc 15 | | | | |
| | | adc 14 | | | | |
| | | adc 14, adc 10 | | | | |
| | | adc 15 | | | | |
| | | adc 2 | | | | |
| | | adc 3 | | | | |
| | | adc 4 | | | | |
| | | adc 5 | | | | |
| | | adc 6 | | | | |
| | | adc 7 | | | | |
| | | adc 8 | | | | |
| | | adc 9 | | | | |
| | | adc 9, adc 5 | | | | |
| | | aa june2011 | | | | |